%% file: arxiv.tex
\documentclass{article}

\usepackage{microtype}
\usepackage{graphicx}
\usepackage{subfigure}
\usepackage{booktabs} 

\usepackage{hyperref}

\usepackage[accepted]{icml2024/icml2024}
\usepackage{amsmath}
\usepackage{amssymb}
\usepackage{mathtools}
\usepackage{amsthm}
\usepackage{algpseudocode}
\usepackage{setspace}
\usepackage{xcolor}
\usepackage[inline]{enumitem}

\MakeRobust{\Call}

\usepackage[capitalize,noabbrev]{cleveref}

\theoremstyle{plain}

\theoremstyle{definition}

\theoremstyle{remark}

\DeclareMathOperator*{\argmin}{arg\,min}

\usepackage[textsize=tiny]{todonotes}

\icmltitlerunning{NoMAD-Attention: Efficient LLM Inference Through Multiply-add-free Attention}

\begin{document}

\twocolumn[
\icmltitle{NoMAD-Attention: Efficient LLM Inference on CPUs Through Multiply-add-free Attention}



\icmlsetsymbol{equal}{*}

\begin{icmlauthorlist}
\icmlauthor{Tianyi Zhang}{rice}
\icmlauthor{Jonah Wonkyu Yi}{rice}
\icmlauthor{Bowen Yao}{rice}
\icmlauthor{Zhaozhuo Xu}{stevens}
\icmlauthor{Anshumali Shrivastava}{rice,thirdai}
\end{icmlauthorlist}

\icmlaffiliation{rice}{Department of Computer Science, Rice University, Houston, TX, USA}
\icmlaffiliation{stevens}{Department of Computer Science, Stevens Institute of Technology, Hoboken, NJ, USA}
\icmlaffiliation{thirdai}{ThirdAI Corp., Houston, TX, USA}

\icmlcorrespondingauthor{Tianyi Zhang}{tz21@rice.edu}
\icmlcorrespondingauthor{Anshumali Shrivastava}{anshumali@rice.edu}

\icmlkeywords{Machine Learning, ICML}

\vskip 0.3in
]



\printAffiliationsAndNotice{} 

\begin{abstract}

Large language model inference on Central Processing Units (CPU) is challenging due to the vast quantities of expensive Multiply-Add (MAD) matrix operations in the attention computations. In this paper, we argue that there is a rare gem in modern CPUs, Single-Instruction-Multiple-Data (SIMD) registers, which allow for ultra-low-latency lookups in batch. We leverage this unique capability of CPUs to propose NoMAD-Attention, an efficient attention algorithm that replaces MAD operations with in-register lookups. Through hardware-aware algorithmic designs, NoMAD-Attention achieves the computation of attention scores using repeated fast accesses to SIMD registers despite their highly limited sizes. Moreover, NoMAD-Attention works with pre-trained attention-based LLMs without model finetuning. Empirical evaluations demonstrate that NoMAD-Attention maintains the quality of the original LLMs well, and speeds up the 4-bit quantized LLaMA-7B-based model by up to $2 \times$ at 16k context length. Our results are reproducible at \url{https://github.com/tonyzhang617/nomad-dist}.


\end{abstract}

\input{icml2024/introduction}

\input{icml2024/background}



\input{icml2024/method}

\input{icml2024/experiments}
\input{icml2024/related_works}

\input{icml2024/conclusion}

\bibliography{icml2024/ref}
\bibliographystyle{icml2024/icml2024}

\newpage
\input{icml2024/appendix}




\end{document}

%% file: icml2024/introduction.tex
\section{Introduction}

Auto-regressive transformer-based Large Language Models (LLM) have demonstrated remarkable abilities across a wide range of natural language processing tasks including reading comprehension, translation, and question answering \cite{unsupervised_learners}. LLMs exhibit emergent abilities \cite{emergent_abilities} in solving complex tasks without fine-tuning. These capabilities give LLMs immense potential for impactful applications in diverse fields such as medicine \cite{llm_in_med}, law~\cite{xiao2021lawformer}, and robotics \cite{llm_applications}.

\textbf{The Need for Deploying LLM on CPUs.} 
Despite the promising potential of LLMs, their deployment is extremely expensive \cite{lin2023pushing}. Serving LLMs with billion-scale parameters requires specialized hardware such as Nvidia A100 Graphics Processing Units (GPUs) \cite{zhang2023hardware}. However, mainstream personal devices, such as laptops, are predominately equipped with Central Processing Units (CPUs) \cite{sun2019summarizing}. As a result, making LLM-related services accessible to everyone remains a major challenge. Reducing the LLM inference latency on CPUs, beyond doubt, has significant implications for its accessibility and adoption.  


\textbf{Expensive Multiply-add Operations for Attention in LLM Inference.} 
LLM inference on CPUs is compute-bound and the primary computational bottleneck is the calculation of attention scores \cite{han2023hyperattention}. Attention, a mechanism that models token interactions through all-pair dot products, heavily relies on the multiply-add (MAD) kernel on processors. The MAD operation involves computing the product of two numbers and adding that product to an accumulator \cite{sung2023mad}. Within the attention mechanism, MAD plays a crucial role in determining the attention score between tokens and subsequently blending their embeddings based on these scores. The computational cost of attention grows quadratically with the sequence length due to the cumulative MAD operations. Since CPUs have limited parallel cores, they are inefficient for handling highly repetitive and parallel workloads. The extensive MAD operations required by the attention mechanism thus become the primary bottleneck during inference. 

\textbf{Opportunities and Challenges from Modern CPUs: In-Register Lookups.} The memory hierarchy of modern CPUs has undergone significant evolution, introducing a new type of registers optimized for Single-Instruction-Multiple-Data (SIMD) operations. The SIMD registers vary in size, ranging from 128 bits to 512 bits \cite{shin2019accelerating}, and support specialized SIMD instructions for high-throughput parallel processing \cite{zhang2019high}. Nowadays, SIMD registers have become a standard feature in commodity hardware, including laptops and mobile devices \cite{dasika2010mighty}. In this context, in-register lookup refers to the low-latency retrieval of information stored within SIMD registers. Specifically, storing information such as dot-product lookup tables (LUT) within SIMD registers as opposed to cache memory has the potential of accelerating LLM inference \cite{qadc}. Despite these opportunities, the limited size of SIMD registers poses a great challenge to fitting the computational paradigm of existing models.




\textbf{Our Proposal: MAD-Free Attention with In-Register Lookups.}  In this paper, we demonstrate a new approach for speeding up LLM inference by leveraging the unique hardware capability of CPUs. We show how the vast quantities of MAD operations in attention computation can be replaced with in-register lookups to mitigate the quadratic computational bottleneck of LLM inference on CPUs. NoMAD-Attention significantly speeds up LLM inference without sacrificing model quality and is compatible with pre-trained attention-based transformers without finetuning.

We summarize our contributions as follows:
\begin{enumerate}
    \item We identify the extensive MAD operations in attention as the bottleneck of CPU LLM inference and explore the opportunity of mitigating it through replacing MAD in attention with fast in-register lookups.
    \item We introduce NoMAD-Attention, a MAD-free framework of attention computation for pre-trained attention-based LLMs. NoMAD-Attention leverages hardware-aware algorithmic designs to enable accurate and fast in-register lookup-based estimations of query-key dot products despite the limited capacity of SIMD registers. NoMAD-Attention preserves model quality while yielding considerable speedups over MAD-based attention.
    \item Through extensive experiments, we demonstrate that NoMAD-Attention achieves up to $2\times$ speedup on 4-bit quantized LLaMA-7B-based models at a context length of 16k, while maintaining the predictive performance of the original model. 

    
\end{enumerate}

%% file: icml2024/background.tex
\section{LLM Inference on CPUs}
In this section, we introduce the attention mechanism used in LLMs and the key-value (KV) caching technique for avoiding redundant attention computations. We also discuss the CPU memory hierarchy, which serves as the motivation for performing fast in-register lookups.


\subsection{LLM Attention}

Most LLMs are decoder-only attention-based models that are pre-trained on a next token prediction objective. LLMs use masked self-attention, in which the attention output of each token is only dependent on previous tokens and itself, and unaffected by future tokens. Masked self-attention allows LLMs to cache key and value embeddings, avoiding future recomputations. However, this comes at the cost of memory overhead.
The autoregressive generation of LLMs consists of two phases \begin{enumerate*}
    \item \textit{prompt processing}, in which the sequence of token embeddings in the prompt is fed through by the model, and their key-value embeddings are cached by the model,
    \item \textit{decoding}, in which a new token is sampled based on the output embedding of the last token, and the embedding of the new token is fed through the model, the output of which becomes the basis for sampling the next token.
\end{enumerate*}
The decoding process continues until an end-of-sequence token \texttt{<EOS>} is sampled.

At the decoding step $t$, a single-head masked self-attention computes its output in the following way. The embedding of the current token $e^t$ is transformed into key, query, and value embeddings through distinct transformations,
\begin{align*}
    k^t = f_K(e^t), q^t = f_Q(e^t), v^t = f_V(e^t)
\end{align*}
Then, the key and value embedding of the current token are appended to the key and value cache, respectively. The KV cache $K^{t-1}_\mathrm{cache}, V^{t-1}_\mathrm{cache}$ of the step $t-1$ contains the key/value embeddings of all previous tokens, and after appending, the KV cache become
\begin{align*}
    K^{t}_\mathrm{cache} = \begin{bmatrix}
        K^{t-1}_\mathrm{cache} \\
        k^t
    \end{bmatrix} = \begin{bmatrix}
        k^1 \\
        k^2 \\
        \dots \\
        k^t
    \end{bmatrix}, V^{t}_\mathrm{cache} = \begin{bmatrix}
        V^{t-1}_\mathrm{cache} \\
        v^t
    \end{bmatrix} = \begin{bmatrix}
        v^1 \\
        v^2 \\
        \dots \\
        v^t
    \end{bmatrix}
\end{align*}
Finally, the attention output is computed as
\begin{align*}
\mathrm{attention}(e^t) = \mathrm{softmax}\left(\frac{q^t(K_{\mathrm{cache}}^t)^\top}{\sqrt{d}}\right)V_{\mathrm{cache}}^t
\end{align*}
where $d$ is the dimensionality of $q^t$. We will refer to the result of $\mathrm{softmax}(\frac{qK^\top}{\sqrt d})$ as the attention scores since they dictate how much ``attention'' each token pays to other tokens. Computations in the prompt processing phase are similar to the decoding phase, except all the prompt tokens are computed in batch. LLMs use multi-head attention, which transforms the concatenation of the outputs of multiple single-head attentions to form an output embedding.

\textbf{MAD-based Attention.} The attention mechanism models the interaction between tokens by performing all-pair dot products, where each dot product is computed via $d$ Multiply-Add (MAD) operations. Since attention computes the interaction between all pairs of tokens exhaustively, the amount of MAD operations scales quadratically with the sequence length, quickly overwhelming the computing capability of CPUs. CPUs are designed to handle complex workloads with granular control, while GPUs are optimized for processing simple and repetitive tasks in high throughput. Hence the success of attention has largely been fueled by the development of highly parallel throughput-oriented processors such as GPUs \cite{flashattention}.

\begin{algorithm}[tb]
   \caption{Attention Score Computation in LLM}
   \label{alg:attn}
\begin{algorithmic}[1]
   \State {\bfseries Input:} query $q^t$, key $k^t$, key cache $K_\mathrm{cache}^{t-1}$
   \State let $K_\mathrm{cache}^t \gets \begin{bmatrix}
K_\mathrm{cache}^{t-1} \\
k^t
\end{bmatrix}$ \Comment{\scriptsize Append the current key to key cache\normalsize}
    \State return $\mathrm{softmax}(\frac{q^t (K_\mathrm{cache}^{t})^\top}{\sqrt{d}})$
\end{algorithmic}
\end{algorithm}

\textbf{MAD-based Attention as Bottleneck of LLM Inference.} The computation of attention scores becomes the bottleneck of LLM inference as the sequence length increases. At the $t$-th step of the decoding phase, the time complexity of computing attention score with MAD is $O(t)$ due to $t$ dot products, while all other components of LLMs such as MLP, skip connections, and normalization have time complexity $O(1)$. We will focus on optimizing the efficiency of attention score computations in our proposed approach. Algorithm \ref{alg:attn} presents the pseudocode for attention score computation, including key caching, for a single-head masked self-attention in LLM. This algorithm will serve as a point of comparison in our proposed approach.

\subsection{Memory Hierarchy of Modern CPUs}

Memory of the CPU is organized into a pyramidal hierarchy as shown in Figure \ref{fig:nomad_attn}, with faster memory being significantly smaller than slower memory. The memory unit with the fastest access speed is registers. Each compute core can access its dedicated registers in just 1-2 CPU cycles, but these registers are highly limited in size, usually not exceeding 64 bits. Modern processors have a new type of registers optimized for Single-Instruction-Multiple-Data (SIMD) operations. These SIMD registers range from 128 bits to 512 bits in size and support specialized SIMD instructions for throughput-oriented parallel processing. SIMD registers are common on commodity hardware, including laptops and mobile devices. Using SIMD operations can speed up deep learning models on CPUs by parallelizing matrix multiplications. However, due to the limited number of cores in a CPU, its efficiency in deep learning is still considerably worse than GPU. Prior works \cite{hashing_dl, slide} have resorted to sparsity and sampling-based approaches to reduce the number of computations for efficient deep learning on CPU, but they require training models from scratch and may not apply to all architectures. In this work, we exploit the SIMD registers to shift the computation paradigm from MAD to in-register lookups, which demonstrates significant speedup over MAD-based models.

When describing our proposed algorithm, we assume SIMD registers are 128 bits wide. There exist systems with wider SIMD registers that support more parallelism, e.g. 256-bit registers in AVX-2 and 512-bit registers in AVX-512. However, the most universal form of SIMD registers uses 128 bits, which is supported by Arm NEON and AVX-compatible processors.

%% file: icml2024/method.tex
\section{Methodology}

In this section, we describe our proposed approach NoMAD-Attention, which replaces MAD operations with in-register lookups to enable fast attention computations on CPUs. NoMAD utilizes three techniques to enable lookup-based attention:
\begin{enumerate*}
    \item transforming dot product computations to memory lookups through product quantization,
    \item compressing lookup tables into SIMD registers for low-latency access,
    \item reorganizing the memory layout of key cache for batch parallel dot product lookups.
\end{enumerate*}


\begin{figure*}
\begin{center}
\centerline{\includegraphics[width=\linewidth]{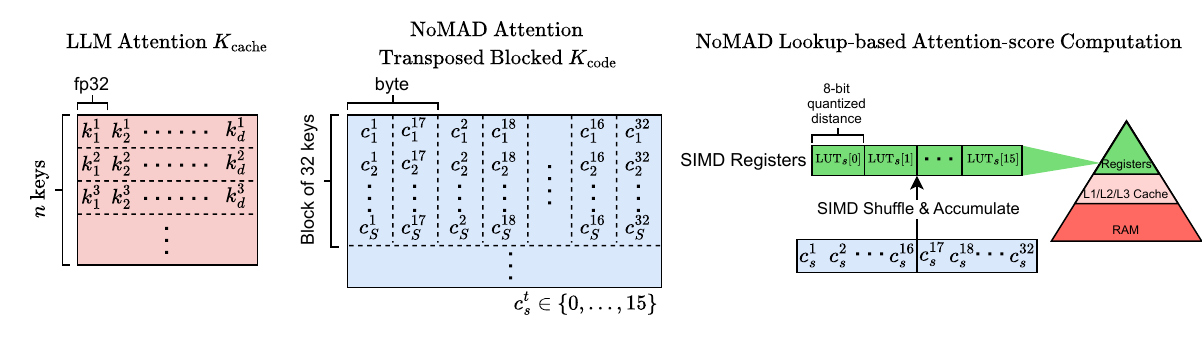}}
\caption{An illustrative comparison of memory layouts of the key cache of LLM attention and the key-code cache of NoMAD-Attention, and an illustration of how attention scores are computed through in-register lookups in NoMAD.}
\label{fig:nomad_attn}
\end{center}
\end{figure*}

\subsection{Transforming Dot-products into Lookups}
Previous works have shown that inexact attention scores in transformers work well for sequence modeling \cite{bigbird}. Based on this insight, NoMAD leverages Product Quantization (PQ) \cite{pq} to compute high-quality estimations of dot products through register lookups. PQ, originally designed for compressing high-dimensional vectors to enable efficient nearest-neighbor search, quantizes a floating-point vector into discrete codes. It makes use of \textit{sub-quantizers}; for a $d$-dimensional vector space, a vector is divided evenly in dimension into $S$ sub-vectors, where each sub-vector has dimension $d_\mathrm{sub} = \frac dS$, and each sub-vector space is quantized independently. We use $\pi_s(e)$, where $s\in \{1\dots S\}$, to denote the function that maps a $d$-dimensional vector $e$ to its $d_\mathrm{sub}$-dimensional sub-vector of the $s$-th sub-quantizer. \textit{Codebooks} are used to quantize sub-vectors to codes, which are collections of cluster centroids learned from a set of training vectors. We use $b_{s,c}$ to denote the $c$-th centroid in the codebook of the $s$-th sub-quantizer. For a given vector $e$, the product-quantized codes of $e$, denoted $c_1, \dots, c_S$, are the indexes of the nearest centroid of each sub-quantizer, i.e.,

\begin{align*}
    \mathrm{PQ}(e) =
    [c_1 \dots c_S],
    \text{where } c_s = \argmin_c \big\lVert \pi_s(v) - b_{s, c}\big\rVert
\end{align*}

Once base vectors have been product-quantized to codes, PQ leverages asymmetric distance computation to keep the estimation error low. In the computed distances, the original query vector is used while the quantized base vectors are used, hence the asymmetry. For a given query $q$, the distances to the centroids of each sub-quantizer are computed and stored in a lookup table (LUT). Then, the corresponding distances in the LUT are looked up based on the codes of base vectors, and accumulated to produce the final distance estimation. Specifically, denoting the distance between query $q$ and the $c$-th centroid for the $s$-th sub-quantizer using $\mathrm{LUT}_s[c] = \mathrm{dist}\big(\pi_s(q), b_{s,c}\big)$, then the estimated distance between query $q$ and a product-quantized base vector $e$, where $
    \mathrm{PQ}(e) = [c_1 \dots c_S]$, is

\begin{equation*}
    \widetilde{\mathrm{dist}}(q, e) = \sum_{s=1}^S \mathrm{LUT}_s[c_s]
\end{equation*}

We extend PQ, which works for metric distances, to estimate dot products for attention. We propose to product-quantize the key vectors in attention to produce key codes, which will be stored in place of the key cache in LLM attention. The codebooks are learned by performing clustering on a set of key vectors from a training set. The key vectors are quantized to the nearest centroid with respect to L2 distance. For a given query, the query-dependent LUT is computed to hold dot products with respect to centroids. Dot products of sub-vectors are retrieved from the LUT based on key codes and accumulated to produce the final dot product estimates. This procedure allows us to compute attention scores through lookups.





\subsection{Compressing Lookup Tables into SIMD Registers}

Estimating dot products through PQ mostly eliminates the use of MAD kernels in the computation of attention scores. However, this approach yields limited speedup over dot-product attention since a high proportion of the CPU cycles are wasted due to cache/memory access stalling (Figure \ref{fig:ablation} offers a comparison between the speed of PQ and dot product operations). It has been shown that even L1-cache-resident LUT is not enough to offer high-performance PQ \cite{fastscan}. The full potential of lookup-based attention can only be unlocked by having the LUT stored in registers, which take only 1-2 CPU cycles to access. However, the highly limited size of registers poses a challenge to fitting the LUT. In PQ, each sub-quantizer commonly uses 256 centroids, which translates to 8-bit codes. Combined with 32-bit floating-point (FP32) dot products, the LUT for each sub-quantizer consumes 8192 bits of memory while the SIMD registers are only 128 bits wide. To circumvent this limitation in register size, we leverage hardware-aware techniques proposed by \citet{qadc} to enable low-latency retrieval from register-resident LUT. 

\textbf{8-bit Quantized Dot Products in LUT} Due to the mere 128-bit width of SIMD registers, the FP32 representation of dot product is too costly to store. Adopting FP32 dot products in LUT implies that each codebook can only contain up to 4 centroids, which will no doubt lead to significant quantization errors. Therefore, we adopt the 8-bit dynamically quantized representation of dot products. Compressing beyond 8-bit is infeasible since many SIMD instruction sets do not support parallel lookups below 8 bits. The quantization is done dynamically for each query to minimize quantization errors. For a given query and sub-quantizer, dot products to centroids are first computed in full FP32 precision. Then the quantization range is determined by the minimum and maximum dot products to the centroids. Finally, the range is evenly divided into $2^8$ buckets and dot products are quantized to the bucket they fall into. More formally, suppose $\mathrm{dp}_{\min} = \min_c (\pi_s(q) \cdot b_{s,c})$ and $\mathrm{dp}_{\max} = \max_c (\pi_s(q) \cdot b_{s,c})$ are the minimum and maximum dot products of the query $q$ to the centroids of the $s$-th sub-quantizer, then the LUT stores the quantized dot products to centroid $c$ as

\begin{equation}
\label{quantize}
    \mathrm{LUT}_s[c] = \Big\lfloor \frac{(\pi_s(q) \cdot b_{s, c}) - \mathrm{dp}_{\min}}{(\mathrm{dp}_{\max}-\mathrm{dp}_{\min})/2^8}\Big\rfloor
\end{equation}

The quantization and de-quantization process can be done efficiently without much computational overhead, and the quantization error is kept low thanks to dynamic query-dependent quantization. 

\textbf{Constrained Codebook Size} By adopting 8-bit quantized dot products in LUT, we can fit 16 dot products on 128-bit SIMD registers. This implies that the codebook size of each sub-quantizer is constrained to 16 centroids. Although the codebook seems limited in size, some evidences suggest it may work well with attention. It has been shown that the output of attention loses rank extremely quickly \cite{attention_loses_rank}, implying that the intermediate embeddings of transformers may exhibit clear clustering structures.



\begin{algorithm}[t]
\setstretch{1.25}
   \caption{\small NoMAD-Attention Score Computation}
   \label{alg:nomad_attn}
\begin{algorithmic}[1]
\small
   \State {\bfseries Input:} query $q^t$, key $k^t$, key-code cache $K_\mathrm{code}^{t-1}$
   \State let $c_s^t \gets \argmin_{c \in \{0, \dots, 15\}} \lVert \pi_s(k^t), b_{s,c}\rVert_2$ for $s=1 \dots S$
   \Statex \Comment{\scriptsize Compute codes for the current key \small}
   \State let $K_\mathrm{code}^t \gets \text{ insert } c_s^t \text{ into } K_\mathrm{code}^{t-1} \text{ for } s = 1 \dots S$
   \Statex \Comment{\scriptsize Insert codes of the current key into the key-code cache \small}
   \State let $\mathrm{LUT}_{s}[c] \gets \mathrm{quantize}(\pi_s(q^t) \cdot b_{s,c})$ {for $s = 1 \dots S, c = 0 \dots 15$}
   \Statex \Comment{\scriptsize Store 8-bit quantized dot products (Equation \ref{quantize}) in LUT\small}
   \State let $\mathrm{accu}[1\dots t] \gets 0$ \Comment{\scriptsize Initialize accumulators\small}
   \For{$i \gets 1 \dots \lceil \frac{t}{32} \rceil$} \Comment{\scriptsize Perform in-register lookups in batch of 32 keys\small}
   \For{$s \gets 1 \dots S$}
    \State \texttt{simd\textunderscore load}{$(\mathrm{LUT}_s)$} \Comment{\scriptsize Load LUT into registers\small}
    \State $\mathrm{accu}[32i-31\dots 32i] \gets $ \texttt{simd\textunderscore add}\big(
    \Statex \hskip 4em {$\mathrm{accu}[32i-31\dots 32i]$,
    \Statex \hskip 4em \texttt{simd\textunderscore shuffle}{$(\mathrm{LUT}_s, K_\mathrm{code}^{32i-31\dots 32i,s})$}}\big)
    \EndFor
   \EndFor
    \State return $\mathrm{softmax}(\frac{\mathrm{dequantize}(\mathrm{accu}[1\dots t])}{\sqrt{d}})$
\end{algorithmic}
\end{algorithm}

\subsection{Reorganizing Key Cache Memory Layout}

Quantized dot products and constrained codebooks enable LUT to be stored in SIMD registers, but the layout format of the key cache needs to be reorganized to take advantage of SIMD instructions. The original key cache in LLM attention stores each key vector contiguously in a row to optimize single vector reads. NoMAD-Attention uses the key-code cache in place of the key cache, which stores the quantized codes of keys. To allow fast lookups of LUT entries based on key codes, we store the key codes in a transposed blocked format. An illustrative comparison between the LLM key cache and the NoMAD key-code cache is given in Figure \ref{fig:nomad_attn}.

The storage format of the NoMAD key-code cache is \textit{transposed}: stored in column-major order instead of row-major, and \textit{blocked}: with 32 keys as a block. The SIMD instruction $\texttt{shuffle}$, which we leverage for performing low-latency batch lookups, takes a batch of byte-size integers as input and retrieves the values held in the registers corresponding to the integer indices. The original storage format of the key cache stores all dimensions of a key contiguously, which does not allow efficient use of $\texttt{shuffle}$. To maximize the usage of the LUT held in registers, we store key codes belonging to the same sub-quantizer contiguously in rows of 32 codes. Since $\texttt{shuffle}$ performs lookups in a batch size of 16, the keys within the same block are stored in alternating order. This is because each quantized code occupies half a byte as there are 16 centroids in a codebook, while the $\texttt{shuffle}$ instruction uses each byte as an input argument. By performing SIMD bit-shifting and bit-masking on a block of alternating keys, we obtain the key codes in the original order, ready for use with $\texttt{shuffle}$. More details on how \texttt{shuffle} is performed on each block of key-code cache and pseudocode can be found in the appendix.

\subsection{NoMAD-Attention}

By combining these three techniques, NoMAD-Attention achieves fast MAD-free attention score computations through SIMD in-register lookups. For a given query, first, LUTs with 8-bit quantized dot products are computed for each sub-quantizer. Then, a LUT is loaded into registers, followed by SIMD \texttt{shuffle} instructions to retrieve dot products in the LUT in batch based on key codes. The loading and lookup are repeated for all sub-quantizers, and the retrieved dot products are accumulated in batch through SIMD \texttt{add}. Finally, the quantized dot products accumulated over all sub-quantizers are de-quantized, scaled, and fed through softmax to produce the attention scores. The pseudocode for NoMAD-Attention score computations is given in Algorithm \ref{alg:nomad_attn}. 




\begin{figure}[t]
\begin{center}
\centerline{\includegraphics[width=\columnwidth]{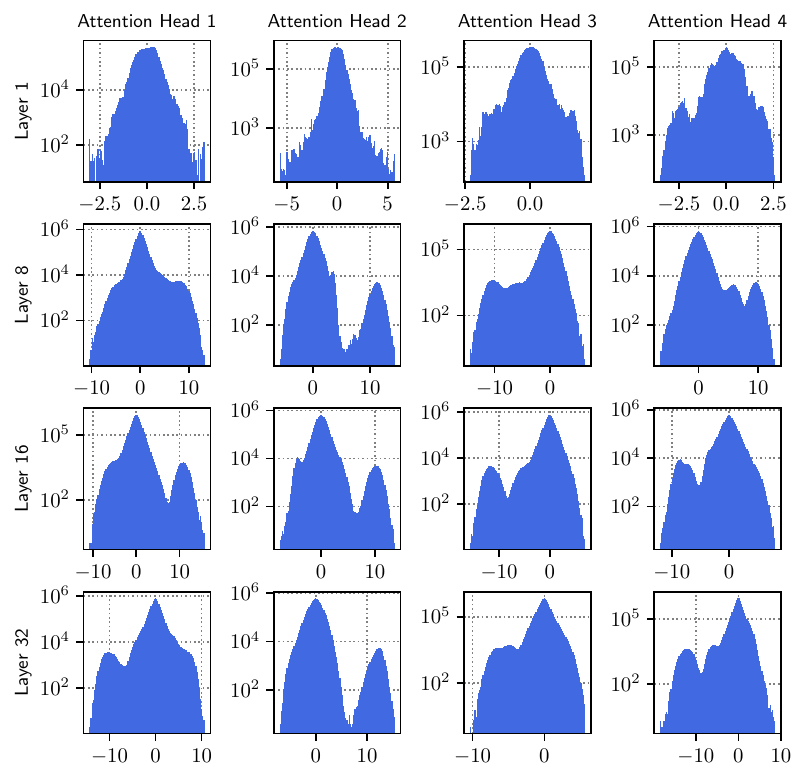}}
\caption{Value distributions of attention key embeddings of the LLaMA-2-7B model on samples of the WikiText-2 dataset. The first 4 attention heads in 4 different layers are shown, and all 128 dimensions of the key embeddings are used. Key embeddings have different distributions in value across different layers and heads, making it necessary for codebooks to be learned independently for each layer and head to minimize quantization error.}
\label{fig:key_dist}
\end{center}
\end{figure}

\begin{figure}[t]
\begin{center}
\centerline{\includegraphics[width=\columnwidth]{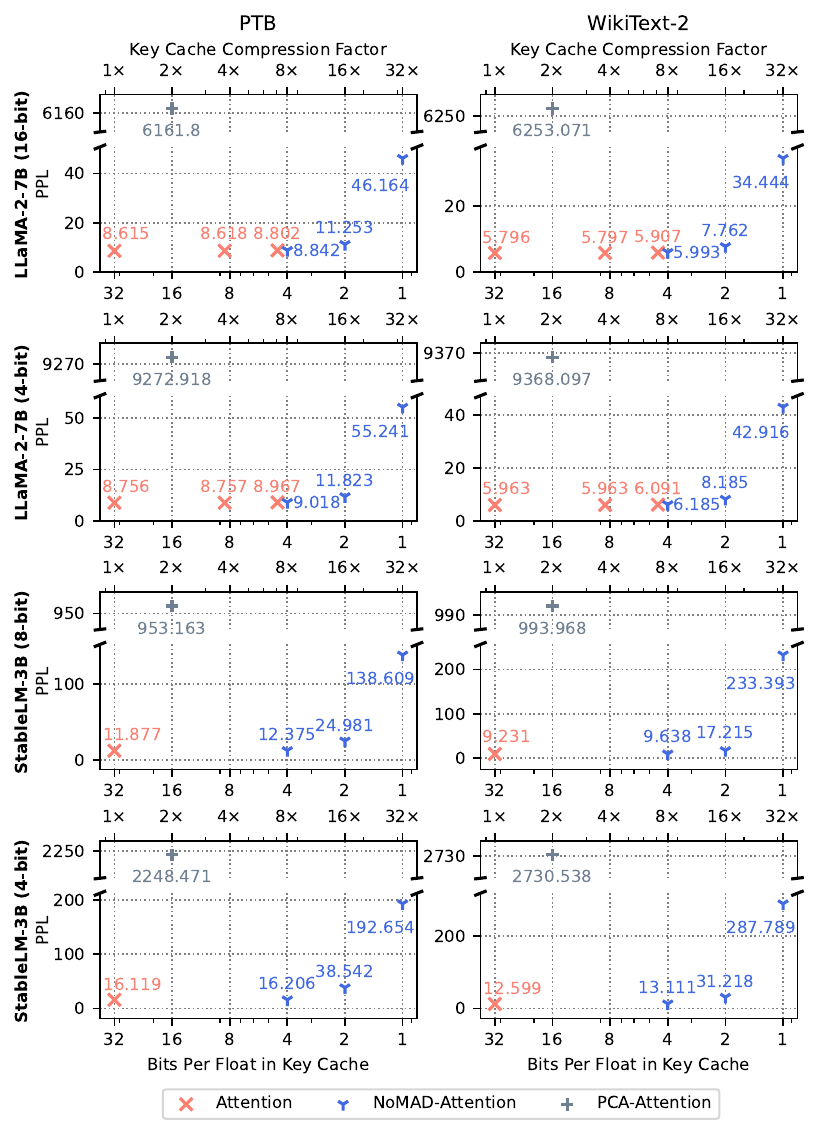}}
\caption{NoMAD-Attention-based LLMs maintain model quality with negligible degradation in perplexity compared to the original model at $8\times$ key cache compression / 4 bits per float in key / $d_\mathrm{sub}=1$. Dimensionality reduction-based PCA-Attention leads to significant model quality degradation even at $2\times$ key cache compression.}
\label{fig:ppl}
\end{center}
\end{figure}

\subsection{Learning Key Compression}

Compressing each segment of the attention key into a 4-bit code requires careful initialization of centroids to avoid high quantization errors, which can lead to degradation in model quality. Ideally, centroids in the codebooks should have low L2 distance to the attention key sub-vectors of the corresponding sub-quantizer. Empirically, we observe that the value distributions in attention key embeddings vary significantly for each attention head and layer. Figure \ref{fig:key_dist} illustrates the value distributions in attention key embeddings of the LLaMA-2-7B model on samples from the WikiText-2 dataset. Distinct attention heads have different value ranges and distributional skew. Hence, we propose learning codebooks for key compression in the following way: we first perform LLM inference with the original attention on a learning set of data and record the attention key embeddings for each layer and head. Subsequently, we learn codebook centroids for key compression by clustering the key embeddings through a K-Means-based algorithm on each attention head independently. These centroids become the codebooks for compressing the key-code cache.






%% file: icml2024/experiments.tex
\section{Experiments}

In this section, we evaluate the effectiveness of our proposed NoMAD-Attention in maintaining model quality and achieving efficient LLM inference on CPUs. In particular, we aim to evaluate \begin{enumerate*}
    \item the model quality of NoMAD-Attention-based LLMs compared to the original LLMs,
    \item the efficiency of NoMAD-Attention-based LLMs compared to attention-based LLMs.
\end{enumerate*}
We first introduce the software implementation and testbed hardware, then detail the experiment setup and baseline methods, and finally report experimental results.

\textbf{Software Implementation} The software system is built in C and C++, based on the open-source projects \texttt{llama.cpp} \footnote{\url{https://github.com/ggerganov/llama.cpp}} and FAISS \cite{faiss}.

\textbf{Testbed Hardware} Experiments are performed on a server running Linux Ubuntu 20.04, equipped with $2\times$ Intel Xeon E5-2695 V3 14-core CPUs, 512GB of DDR4 RAM, and 1TB of SSD. The processors used support AVX2 SIMD instructions, which we leverage to perform in-register lookups for NoMAD-Attention.

\begin{figure*}[]
\begin{center}
\centerline{\includegraphics[width=\linewidth]{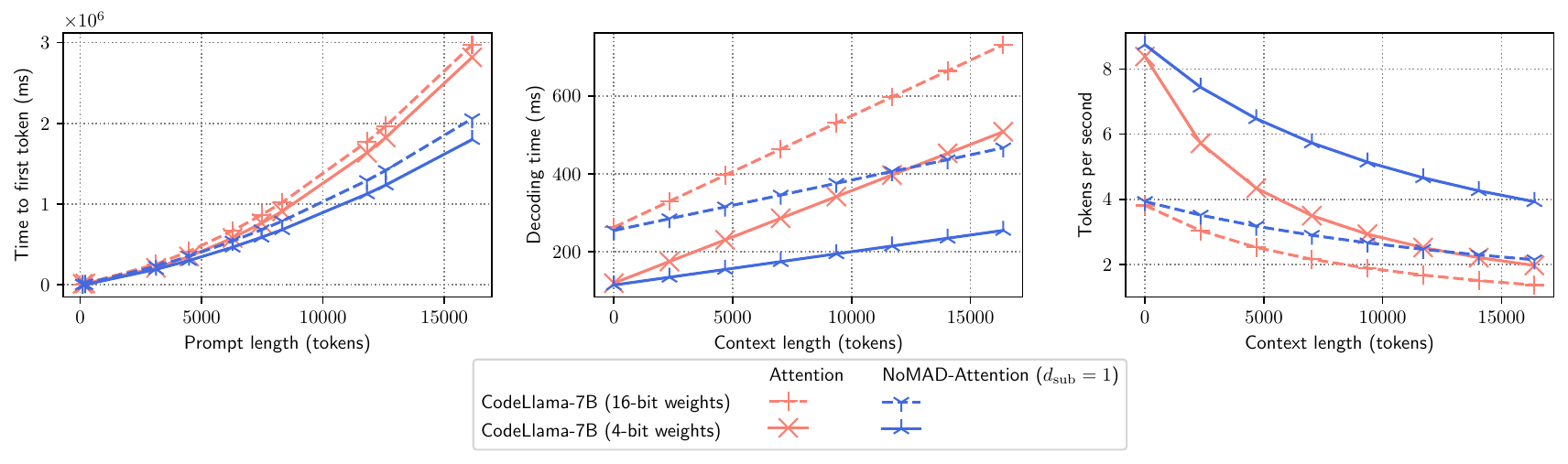}}
\caption{The efficiency of Attention-based and NoMAD-Attention-based CodeLLaMA-7B models on prompt processing and decoding. NoMAD-Attention-based models achieve significant speedup over Attention-based counterparts. At the context length of 16k, NoMAD-Attention-based CodeLlama-7B (4-bit weights) achieves $2\times$ speedup over the original CodeLlama-7B (4-bit weights).}
\label{fig:efficiency}
\end{center}
\end{figure*}

\subsection{Experiment Setup}

\subsubsection{Measuring LLM Quality}
We first perform a set of experiments to measure the model quality of NoMAD-Attention-based LLMs against baselines. Model quality is measured using the perplexity metric (lower the better), which is defined as
\begin{align*}
    \mathrm{PPL}(x) = \exp\left(-\frac 1T\sum_{i=1}^T\log P(x_i | x_{<i})\right)
\end{align*}
where $x = \{x_i\}_{1 \le i \le T}$ is a sequence of target tokens, and $P(x_i | x_{<i})$ is the probability of token $x_i$ being predicted by the model conditioned on the previous tokens $x_{<i}$ as context. Perplexity is measured on the test set of two datasets, WikiText-2 \cite{wikitext2} and Penn Treebank (PTB) \cite{ptb}, in chunks of length 512. The LLMs employed in perplexity testing are LLaMA-2-7B \cite{llama2} (with the original 16-bit and quantized 4-bit weights) and StableLM-3B-4E1T \cite{StableLM-3B} (with 8-bit and 4-bit quantized weights).



\textbf{Baselines} We use LLMs with the original dot-product attention as a baseline, and evaluate the model quality of NoMAD-Attention-based LLMs and PCA-Attention-based LLMs. Principal Component Analysis (PCA) \cite{wold1987principal} is a well-used and studied dimensionality reduction technique. By reducing the dimensionality of query and key embeddings via PCA, the efficiency of attention score computation can be improved. NoMAD and PCA require codebook learning and projection learning, respectively, in which we use the key embeddings of the first 100 samples from the training set of WikiText-2 and PTB datasets for learning. This avoids the train-test overlap and ensures that the codebooks learned can generalize to unseen data. For the same model, the codebooks and projection are learned only once and used for different quantization schemes. For the original dot-product attention, we vary the compression of the key cache from FP32 to \texttt{q4\textunderscore0} and \texttt{q8\textunderscore0} (each float uses 4.5 and 8.5 bits respectively). For StableLM models, \texttt{q4\textunderscore0} and \texttt{q8\textunderscore0} key cache compression is not supported by the \texttt{llama.cpp} library, hence omitted. For NoMAD and PCA, all attention heads in the LLM are replaced with their attention variant.

\subsubsection{Measuring LLM Efficiency}

For measuring the model efficiency, we use CodeLlama-7B \cite{codellama} (with 16-bit and 4-bit weights), a variant of the Llama LLM that supports a long context length of 16384. We sample 10 sequences of varying lengths up to 16K from the stack-overflow-questions dataset \cite{stackoverflow_questions}, and use them as prompts to generate 4096 tokens. The baseline implementation is based on the \texttt{llama.cpp} implementation. The experiments are run with all available 28 CPU cores. For efficiency comparisons, we report the time to the first token (time to finish prompt processing), decoding time for each token, and decoded tokens per second.

\subsection{LLM Quality}
The results of the model quality comparison are shown in Figure \ref{fig:ppl}. Overall, the NoMAD-Attention-based LLM maintains model quality at $8\times$ key cache compression ($d_\mathrm{sub}=1$) with a negligible loss in perplexity (consistently less than a 4\% increase). In contrast, the dimensionality-reduction-based method PCA fails to maintain model quality at $2\times$ key cache compression. At $8\times$ key cache compression, or $d_\mathrm{sub}=1$, NoMAD-Attention-based LLMs maintain model quality well compared to the original Attention-based LLMs, as demonstrated by the perplexity metric. Beyond $8\times$ key cache compression, NoMAD-Attention-based LLMs drop in quality with increasing compression factor. However, NoMAD-Attention is significantly better than PCA-Attention in maintaining model quality. The quality of attention drops catastrophically when dimensionality reduction is applied. This is likely because dimensionality reduction strategies use symmetric dot-product computations, while NoMAD uses asymmetric dot-product computations.

\subsection{LLM Efficiency}

The experimental results of the model efficiency comparison are given in Figure \ref{fig:efficiency}. Since NoMAD-Attention maintains model quality well at $8\times$ key cache compression or $d_\mathrm{sub}=1$, we explore the speedup of NoMAD-Attention-based models at this configuration. We also include results of NoMAD-Attention at $d_\mathrm{sub}=2$ in the appendix. NoMAD-Attention-based LLMs achieve significant speedups over the original models. We highlight that NoMAD-Attention-based CodeLlama-7B (4-bit weights) achieves $2\times$ speedup over the original CodeLlama-7B (4-bit weights) at 16k sequence length.

\subsection{Ablation Study}

\begin{figure}
\begin{center}
\centerline{\includegraphics[width=\columnwidth]{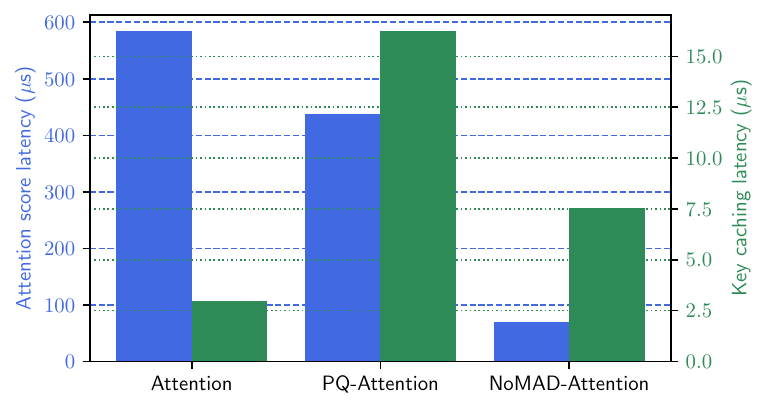}}
\caption{The latency per query of Attention, PQ-Attention (8-bit code, $d_\mathrm{sub}=2$), and NoMAD-Attention (4-bit code, $d_\mathrm{sub}=1$) in computing attention scores and key caching for 16k queries at 16k context length. PQ-Attention yields limited speedup compared to Attention and incur the most overhead in key caching due to the large size of codebooks. NoMAD-Attention significantly reduces the latency of attention score computations over Attention.}
\label{fig:ablation}
\end{center}
\end{figure}

We study the efficiency of attention score computation and key caching for single-head attention, NoMAD-Attention, and PQ-Attention to study the effectiveness of our proposed hardware-aware strategies. PQ-Attention quantizes keys into 8-bit codes in each sub-quantizer and performs asymmetric dot product computations. To match the key compression factor, we use PQ-Attention at $d_\mathrm{sub}=2$ and NoMAD-Attention at $d_\mathrm{sub}=1$. We perform 16k queries at a context length of 16k and measure the latency of each attention in attention score computation and key caching using a single thread. The results of the ablation study are given in Figure \ref{fig:ablation}. Despite replacing MADs with lookups, PQ-Attention yields limited speedup as compared to dot-product attention. NoMAD-Attention achieves $8.3\times$ speedup over MAD-based attention. The key caching time of PQ-Attention is $2\times$ more than NoMAD-Attention, since finding the code for each sub-quantizer takes 256 distance computations as opposed to 16.

%% file: icml2024/related_works.tex
\section{Related Works}

\paragraph{Efficient and Approximate Attention} Since the introduction of attention in transformers \cite{attention}, there has been a body of work on approximating the attention mechanism for efficient training and inference of transformers. For example, dynamically sparse attention has been achieved using LSH \cite{reformer}, Nystr\"om method \cite{nystromformer}, and random sampling \cite{bigbird}. Furthermore, low-rank attention has been extensively explored \cite{linformer, performer, scatterbrain} and shown to have compute- and memory-efficiency advantages over regular transformers. Attention mechanisms with hardware-aware designs such as FlashAttention \cite{flashattention} have been proposed to mitigate the IO bottleneck in GPUs. In large language models, multiple approaches \cite{h2o, scissorhands} have been proposed to reduce the high memory overhead of the KV cache. For CPU-only environments, \citet{llm_cpu} proposes to speed up LLM inference through weight quantization.
\vspace{-10 pt}
\paragraph{Matrix Multiplication Optimization and Compression} Approximate matrix multiplication is applicable in a wide range of computational problems from statistical analysis to image compression, and its optimization has been a topic of interest for years \cite{compressedmatrixmultiplication}. Using novel compression techniques and specialized hardware, modern researchers have begun optimizing matrix multiplication around the specific limitations of computers including their memory capacity and traffic between the CPU and main memory \cite{onlargescalematrixmatrix}. Compression techniques were developed to multiply billion-scale matrices, fully utilize the DSP, and use learning-based algorithms on computers \cite{billionscalematrixcompression, multiplyingmatriceswithout, onlargescalematrixmatrix}. However, many of these algorithms still had limitations ranging from training on a matrix to megabytes of hardware resources and computation that made the matrix multiplication of LLMs nearly impossible to compute without prior training or significant hardware resources \cite{multiplyingmatriceswithout, onlargescalematrixmatrix}.

%% file: icml2024/conclusion.tex
\section{Conclusion}

In conclusion, this study aims to address the challenges of large language model inference on Central Processing Units (CPUs), particularly the difficulties associated with the expensive Multiply-Add (MAD) matrix operations in attention mechanisms. The investigation highlighted the untapped potential of Single-Instruction-Multiple-Data (SIMD) registers and their fast in-register lookup capabilities within CPUs. The proposed NoMAD-Attention algorithm was introduced as an efficient alternative to traditional MAD-based approaches, leveraging in-register lookups and optimizing memory access to SIMD registers. Consequently, the implementation of NoMAD-Attention resulted in a significant acceleration of LLaMA-7B-based model inference, achieving up to a $2\times$ speedup on CPUs. This research underscores the importance of exploring novel approaches, such as NoMAD-Attention, to enhance the efficiency of large language model inference on CPU architectures.

\section{Impact Statement}
This paper aims to democratize large language models (LLMs) by enabling their operation on CPU cores, making them accessible to a broader audience. By successfully demonstrating the implementation of an LLM on CPU, our study contributes to fostering innovation and expansion of cutting-edge LLM technologies to a wider user base.

%% file: icml2024/appendix.tex
\appendix
\onecolumn
\section*{Appendix}

\section{Details Regarding SIMD Instructions}

The SIMD $\texttt{shuffle}$ presented in Algorithm \ref{alg:nomad_attn} is a simplification of the actual hardware implementation. We give full details of lines 5 to 11 in Algorithm \ref{alg:nomad_attn} in Algorithm \ref{alg:nomad_attn_loop}. Keys are stored in blocks of 32, in which keys are stored in an alternating order (see Figure \ref{fig:nomad_attn} for an illustration). After a LUT is loaded into registers, the row of key codes in the block corresponding to the sub-quantizer is used to perform $\texttt{shuffle}$. First, each byte in the row is bit-shifted to the right by 4 bits via a SIMD instruction, which produces the codes of the first 16 keys in the block. The codes are fed to $\texttt{shuffle}$ to retrieve the quantized dot products of the first 16 keys from the LUT. Then, the first 4 bits of each byte in the row are masked out via a SIMD instruction, which produces the code of the last 16 keys in the block. They are similarly used to retrieve the quantized dot products from the LUT. The retrieved quantized dot products of 32 keys are accumulated in the accumulator. Since quantized dot products are 8 bits wide, accumulating them in 8-bit accumulators easily results in overflows. Therefore, 16-bit accumulators are used to accumulate quantized dot products.

\begin{algorithm}[h]
\setstretch{1.25}
   \caption{\small NoMAD Dot-Product Lookup Accumulation Loop}
   \label{alg:nomad_attn_loop}
\begin{algorithmic}[1]
\small
   \State let $\mathrm{accu}[1\dots t] \gets 0$ \Comment{\scriptsize Initialize 16-bit unsigned accumulators\small}
   \For{$i \gets 1 \dots \lceil \frac{t}{32} \rceil$}
   \For{$s \gets 1 \dots S$}
    \State \texttt{simd\textunderscore load}({$\mathrm{LUT}_s$}) \Comment{\scriptsize Load LUT into registers\small}
    \State let $K_\mathrm{cache}^{32i-31\dots 32i-16,s} \gets$\texttt{simd\textunderscore bitwise\textunderscore right\textunderscore shift}$(K_\mathrm{cache}^{32i-31\dots 32i,s}, 4)$ \Comment{Obtain the first 15 key codes through bit shifting}
    \State $\mathrm{accu}[32i-31\dots 32i-16] \gets $ \texttt{simd\textunderscore add}\big(
    \Statex \hskip 4em {$\mathrm{accu}[32i-31\dots 32i-16]$,
    \Statex \hskip 4em \texttt{simd\textunderscore shuffle}{$(\mathrm{LUT}_s, K_\mathrm{cache}^{32i-31\dots 32i-16,s})$}}\big)
    \State let $K_\mathrm{cache}^{32i-15\dots 32i,s} \gets$\texttt{simd\textunderscore bitwise\textunderscore and}$(K_\mathrm{cache}^{32i-15\dots 32i,s}, \texttt{0xf})$ \Comment{Obtain the last 15 key codes through bitwise and}
    \State $\mathrm{accu}[32i-15\dots 32i] \gets $ \texttt{simd\textunderscore add}\big(
    \Statex \hskip 4em {$\mathrm{accu}[32i-15\dots 32i]$,
    \Statex \hskip 4em \texttt{simd\textunderscore shuffle}{$(\mathrm{LUT}_s, K_\mathrm{cache}^{32i-15\dots 32i,s})$}}\big)
    \EndFor
   \EndFor
\end{algorithmic}
\end{algorithm}

\begin{figure}[H]
\begin{center}
\centerline{\includegraphics[width=0.5\columnwidth]{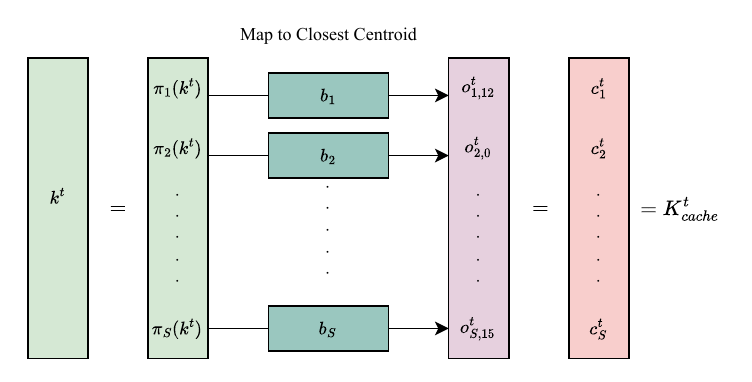}}
\caption{Illustration demonstrating the mapping of an input key $k^t$ to its $s$-th sub-quantizer using $\pi_s(k^t)$, where $s\in {1\dots S}$. Subsequently, each sub-quantizer maps to its closest centroid $c_i^t$, where $i\in {1\dots S}$, and the results are stored in the key cache $K_\mathrm{cache}^t$.}
\label{fig:keyq}
\end{center}
\end{figure}

\begin{figure}[H]
\begin{center}
\centerline{\includegraphics[width=\columnwidth]{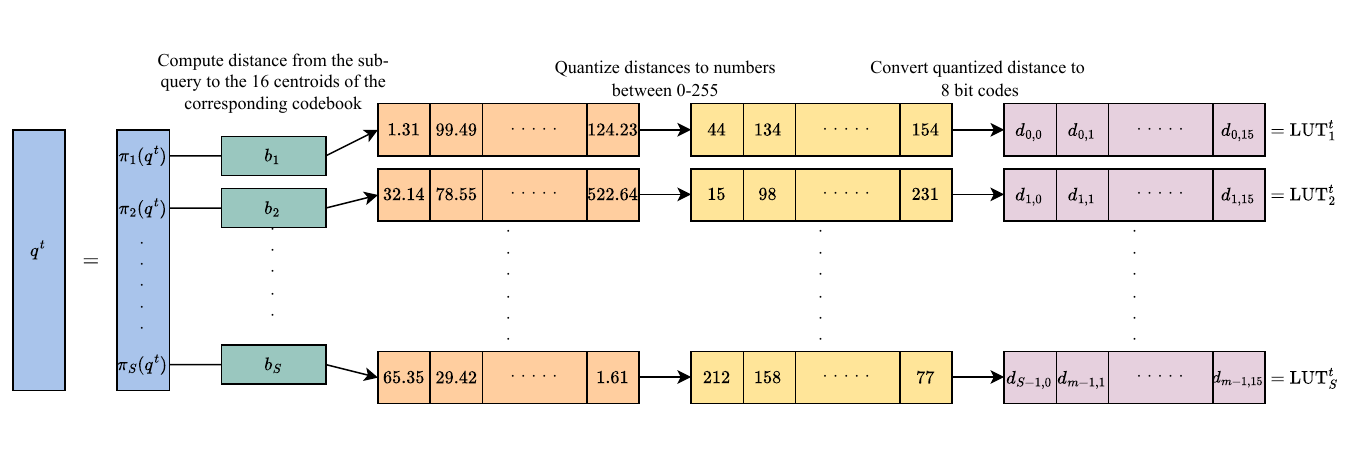}}
\caption{Illustration depicting the mapping of a query vector $q$ to its $s$-th sub-quantizer using $\pi_s(q)$, where $s\in {1\dots S}$. Subsequently, the distance between $\pi_s(q)$ and 16 centroids is computed. This distance is quantized to a value within the range of 0-255, and the resulting quantized distance is further converted into 8-bit codes, which are stored in $LUT_s$.}
\label{fig:queryq}
\end{center}
\end{figure}

\section{Visual Explanations on $\boldsymbol{K_\mathrm{cache}}$ and Lookup Table (LUT) Construction}
Figure \ref{fig:keyq} illustrates the process of mapping and compressing key vector $k^t$ to construct $K_\mathrm{cache}^t$. For an input key vector $k^t$, functions $\pi_s$, where $s\in {1\dots S}$, split the vector into sub-vectors $k^t = (\pi_1(k^t), \pi_2(k^t), \pi_S(k^t))$. Subsequently, each sub-quantizer $\pi_s(k^t)$ is mapped to its nearest centroid $c_s^t$ by referencing the codebook $b_s$, where $i\in {1\dots S}$, among 16 centroids in the codebook. The resulting values are then stored in the key cache $K_\mathrm{cache}^t$.

Similarly, Figure \ref{fig:queryq} illustrates the process of mapping and compressing query vector $q^t$ to construct the Look-up Tables (LUT). Given a query vector $q^t$, functions $\pi_s$, where $s\in {1\dots S}$, first split the query into sub-queries $q = (\pi_1(q^t), \pi_2(q^t), \pi_S(q^t))$. Subsequently, the distances between each sub-query $\pi_s(q^t)$ and the 16 centroids from the codebook $b_s$ are computed and then quantized to values within the range of 0-255. Lastly, the quantized vectors are converted into 8-bit codes and stored in $\mathrm{LUT}_s^t$.

\begin{figure*}[b]
\begin{center}
\centerline{\includegraphics[width=\linewidth]{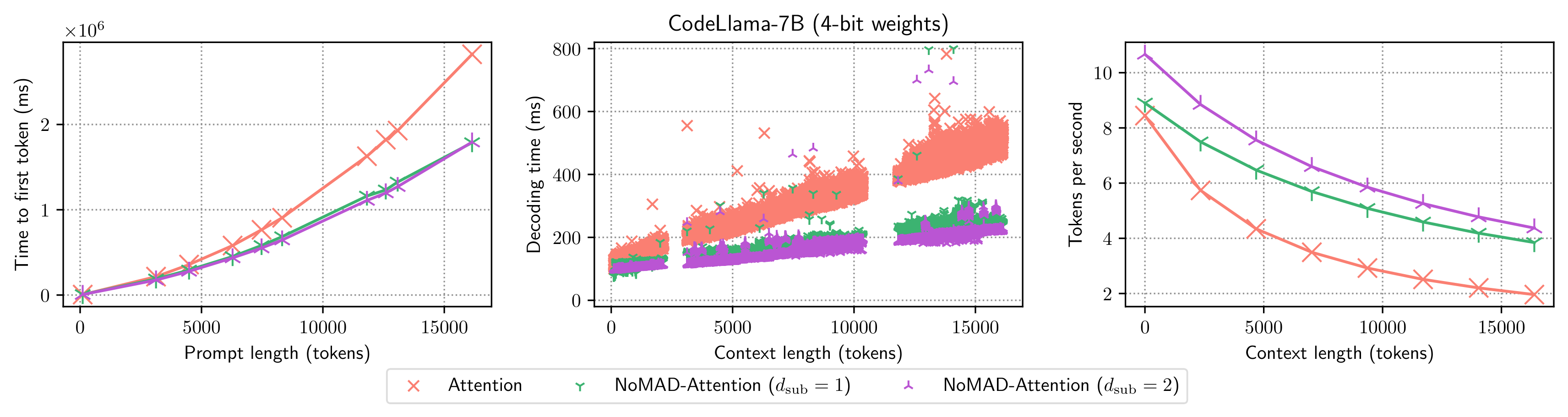}}
\caption{A speed comparison of NoMAD-Attention-based CodeLlama-7B (4-bit quantized weights) at $d_\mathrm{sub}=1$ and $d_\mathrm{sub}=2$ with the original LLM.}
\label{fig:dsub2}
\end{center}
\end{figure*}

\section{Efficiency of NoMAD-Attention at Different Compression Rates}
The results of the model efficiency comparison are depicted in Figure \ref{fig:dsub2}. Concerning the time required to finish prompt processing, the original model with 4-bit quantized weights takes $2.8 \times 10^6$ ms, while the NoMAD-Attention-based CodeLlama-7B only requires approximately $1.8 \times 10^6$ ms for models with both $d_{\text{sub}} = 1$ and $d_{\text{sub}} = 2$, achieving over a $1.5 \times$ increase at a 16k prompt length.

Regarding the decoding time for each token, the original model with 4-bit quantized weights takes $450-600$ ms, while the NoMAD-Attention-based CodeLlama-7B only requires approximately $220$ and $200$ ms for models with $d_{\text{sub}} = 1$ and $d_{\text{sub}} = 2$ respectively, achieving over a $2 \times$ increase at a 16k prompt length.

In terms of throughput measured by tokens per second, at a context length of 16k, the NoMAD-Attention-based CodeLlama-7B can achieve speeds of $4$ and $2.2$ tokens per second on models with 16-bit and 4-bit quantized weights, respectively. In contrast, the original model only manages $2$ and $0.8$ tokens per second, demonstrating more than a $2 \times$ increase at the 16k context length. The original model with 4-bit quantized weights only manages $2$ tokens per second, while the NoMAD-Attention-based CodeLlama-7B can achieve speeds of up to $4$ and $4.2$ tokens per second for models with $d_{\text{sub}} = 1$ and $d_{\text{sub}} = 2$ respectively, also achieving over a $2 \times$ increase at a 16k prompt length.

Overall, the NoMAD-Attention-based CodeLlama-7B (4-bit quantized weights) achieves a $2\times$ speedup over the original CodeLlama-7B (4-bit quantized weights) at long prompt and context lengths (e.g., 16k).

%% file: arxiv.bbl
\begin{thebibliography}{42}
\providecommand{\natexlab}[1]{#1}
\providecommand{\url}[1]{\texttt{#1}}
\expandafter\ifx\csname urlstyle\endcsname\relax
  \providecommand{\doi}[1]{doi: #1}\else
  \providecommand{\doi}{doi: \begingroup \urlstyle{rm}\Url}\fi

\bibitem[Andr{\'e} et~al.(2016)Andr{\'e}, Kermarrec, and Le~Scouarnec]{fastscan}
Andr{\'e}, F., Kermarrec, A.-M., and Le~Scouarnec, N.
\newblock Cache locality is not enough: High-performance nearest neighbor search with product quantization fast scan.
\newblock In \emph{42nd International Conference on Very Large Data Bases}, volume~9, pp.\ ~12, 2016.

\bibitem[Andr{\'e} et~al.(2017)Andr{\'e}, Kermarrec, and Le~Scouarnec]{qadc}
Andr{\'e}, F., Kermarrec, A.-M., and Le~Scouarnec, N.
\newblock Accelerated nearest neighbor search with quick adc.
\newblock In \emph{Proceedings of the 2017 ACM on International Conference on Multimedia Retrieval}, pp.\  159--166, 2017.

\bibitem[Annamoradnejad et~al.(2022)Annamoradnejad, Habibi, and Fazli]{stackoverflow_questions}
Annamoradnejad, I., Habibi, J., and Fazli, M.
\newblock Multi-view approach to suggest moderation actions in community question answering sites.
\newblock \emph{Information Sciences}, 600:\penalty0 144--154, 2022.
\newblock ISSN 0020-0255.
\newblock \doi{https://doi.org/10.1016/j.ins.2022.03.085}.
\newblock URL \url{https://www.sciencedirect.com/science/article/pii/S0020025522003127}.

\bibitem[Blalock \& Guttag(2021)Blalock and Guttag]{multiplyingmatriceswithout}
Blalock, D. and Guttag, J.
\newblock Multiplying matrices without multiplying.
\newblock In Meila, M. and Zhang, T. (eds.), \emph{Proceedings of the 38th International Conference on Machine Learning}, volume 139 of \emph{Proceedings of Machine Learning Research}, pp.\  992--1004. PMLR, 18--24 Jul 2021.
\newblock URL \url{https://proceedings.mlr.press/v139/blalock21a.html}.

\bibitem[Chen et~al.(2020)Chen, Medini, Farwell, Tai, Shrivastava, et~al.]{slide}
Chen, B., Medini, T., Farwell, J., Tai, C., Shrivastava, A., et~al.
\newblock Slide: In defense of smart algorithms over hardware acceleration for large-scale deep learning systems.
\newblock \emph{Proceedings of Machine Learning and Systems}, 2:\penalty0 291--306, 2020.

\bibitem[Chen et~al.(2021)Chen, Dao, Winsor, Song, Rudra, and R{\'e}]{scatterbrain}
Chen, B., Dao, T., Winsor, E., Song, Z., Rudra, A., and R{\'e}, C.
\newblock Scatterbrain: Unifying sparse and low-rank attention.
\newblock \emph{Advances in Neural Information Processing Systems}, 34:\penalty0 17413--17426, 2021.

\bibitem[Choromanski et~al.(2020)Choromanski, Likhosherstov, Dohan, Song, Gane, Sarlos, Hawkins, Davis, Mohiuddin, Kaiser, et~al.]{performer}
Choromanski, K., Likhosherstov, V., Dohan, D., Song, X., Gane, A., Sarlos, T., Hawkins, P., Davis, J., Mohiuddin, A., Kaiser, L., et~al.
\newblock Rethinking attention with performers.
\newblock \emph{arXiv preprint arXiv:2009.14794}, 2020.

\bibitem[Dao et~al.(2022)Dao, Fu, Ermon, Rudra, and R{\'e}]{flashattention}
Dao, T., Fu, D., Ermon, S., Rudra, A., and R{\'e}, C.
\newblock Flashattention: Fast and memory-efficient exact attention with io-awareness.
\newblock \emph{Advances in Neural Information Processing Systems}, 35:\penalty0 16344--16359, 2022.

\bibitem[Dasika et~al.(2010)Dasika, Woh, Seo, Clark, Mudge, and Mahlke]{dasika2010mighty}
Dasika, G., Woh, M., Seo, S., Clark, N., Mudge, T., and Mahlke, S.
\newblock Mighty-morphing power-simd.
\newblock In \emph{Proceedings of the 2010 international conference on Compilers, architectures and synthesis for embedded systems}, pp.\  67--76, 2010.

\bibitem[Dong et~al.(2021)Dong, Cordonnier, and Loukas]{attention_loses_rank}
Dong, Y., Cordonnier, J.-B., and Loukas, A.
\newblock Attention is not all you need: Pure attention loses rank doubly exponentially with depth.
\newblock In \emph{International Conference on Machine Learning}, pp.\  2793--2803. PMLR, 2021.

\bibitem[Douze et~al.(2024)Douze, Guzhva, Deng, Johnson, Szilvasy, Mazaré, Lomeli, Hosseini, and Jégou]{faiss}
Douze, M., Guzhva, A., Deng, C., Johnson, J., Szilvasy, G., Mazaré, P.-E., Lomeli, M., Hosseini, L., and Jégou, H.
\newblock The faiss library.
\newblock 2024.

\bibitem[Han et~al.(2023)Han, Jayaram, Karbasi, Mirrokni, Woodruff, and Zandieh]{han2023hyperattention}
Han, I., Jayaram, R., Karbasi, A., Mirrokni, V., Woodruff, D.~P., and Zandieh, A.
\newblock Hyperattention: Long-context attention in near-linear time.
\newblock \emph{arXiv preprint arXiv:2310.05869}, 2023.

\bibitem[Jegou et~al.(2010)Jegou, Douze, and Schmid]{pq}
Jegou, H., Douze, M., and Schmid, C.
\newblock Product quantization for nearest neighbor search.
\newblock \emph{IEEE transactions on pattern analysis and machine intelligence}, 33\penalty0 (1):\penalty0 117--128, 2010.

\bibitem[Kaddour et~al.(2023)Kaddour, Harris, Mozes, Bradley, Raileanu, and McHardy]{llm_applications}
Kaddour, J., Harris, J., Mozes, M., Bradley, H., Raileanu, R., and McHardy, R.
\newblock Challenges and applications of large language models.
\newblock \emph{arXiv preprint arXiv:2307.10169}, 2023.

\bibitem[Kitaev et~al.(2020)Kitaev, Kaiser, and Levskaya]{reformer}
Kitaev, N., Kaiser, {\L}., and Levskaya, A.
\newblock Reformer: The efficient transformer.
\newblock \emph{arXiv preprint arXiv:2001.04451}, 2020.

\bibitem[Krishna et~al.(2021)Krishna, Narasimhan, Radhakrishnan, and Veras]{onlargescalematrixmatrix}
Krishna, S.~G., Narasimhan, A., Radhakrishnan, S., and Veras, R.
\newblock On large-scale matrix-matrix multiplication on compressed structures.
\newblock In \emph{2021 IEEE International Conference on Big Data (Big Data)}, pp.\  2976--2985. IEEE, 2021.

\bibitem[Lin et~al.(2023)Lin, Qu, Chen, Chen, Chen, and Huang]{lin2023pushing}
Lin, Z., Qu, G., Chen, Q., Chen, X., Chen, Z., and Huang, K.
\newblock Pushing large language models to the 6g edge: Vision, challenges, and opportunities.
\newblock \emph{arXiv preprint arXiv:2309.16739}, 2023.

\bibitem[Liu et~al.(2023)Liu, Desai, Liao, Wang, Xie, Xu, Kyrillidis, and Shrivastava]{scissorhands}
Liu, Z., Desai, A., Liao, F., Wang, W., Xie, V., Xu, Z., Kyrillidis, A., and Shrivastava, A.
\newblock Scissorhands: Exploiting the persistence of importance hypothesis for llm kv cache compression at test time.
\newblock \emph{arXiv preprint arXiv:2305.17118}, 2023.

\bibitem[Marcus et~al.(1993)Marcus, Santorini, and Marcinkiewicz]{ptb}
Marcus, M.~P., Santorini, B., and Marcinkiewicz, M.~A.
\newblock Building a large annotated corpus of {E}nglish: The {P}enn {T}reebank.
\newblock \emph{Computational Linguistics}, 19\penalty0 (2):\penalty0 313--330, 1993.
\newblock URL \url{https://www.aclweb.org/anthology/J93-2004}.

\bibitem[Merity et~al.(2016)Merity, Xiong, Bradbury, and Socher]{wikitext2}
Merity, S., Xiong, C., Bradbury, J., and Socher, R.
\newblock Pointer sentinel mixture models, 2016.

\bibitem[Nelson et~al.(2019)Nelson, Radhakrishnan, and Sekharan]{billionscalematrixcompression}
Nelson, M., Radhakrishnan, S., and Sekharan, C.~N.
\newblock Billion-scale matrix compression and multiplication with implications in data mining.
\newblock In \emph{2019 IEEE 20th International Conference on Information Reuse and Integration for Data Science (IRI)}, pp.\  395--402. IEEE, 2019.

\bibitem[Pagh(2013)]{compressedmatrixmultiplication}
Pagh, R.
\newblock Compressed matrix multiplication.
\newblock \emph{ACM Transactions on Computation Theory (TOCT)}, 5\penalty0 (3):\penalty0 1--17, 2013.

\bibitem[Radford et~al.(2019)Radford, Wu, Child, Luan, Amodei, Sutskever, et~al.]{unsupervised_learners}
Radford, A., Wu, J., Child, R., Luan, D., Amodei, D., Sutskever, I., et~al.
\newblock Language models are unsupervised multitask learners.
\newblock \emph{OpenAI blog}, 1\penalty0 (8):\penalty0 9, 2019.

\bibitem[Roziere et~al.(2023)Roziere, Gehring, Gloeckle, Sootla, Gat, Tan, Adi, Liu, Remez, Rapin, et~al.]{codellama}
Roziere, B., Gehring, J., Gloeckle, F., Sootla, S., Gat, I., Tan, X.~E., Adi, Y., Liu, J., Remez, T., Rapin, J., et~al.
\newblock Code llama: Open foundation models for code.
\newblock \emph{arXiv preprint arXiv:2308.12950}, 2023.

\bibitem[Shen et~al.(2023)Shen, Chang, Dong, Luo, and Meng]{llm_cpu}
Shen, H., Chang, H., Dong, B., Luo, Y., and Meng, H.
\newblock Efficient llm inference on cpus.
\newblock \emph{arXiv preprint arXiv:2311.00502}, 2023.

\bibitem[Shin et~al.(2019)Shin, Choo, and Park]{shin2019accelerating}
Shin, S.-R., Choo, S.-Y., and Park, J.-S.
\newblock Accelerating random network coding using 512-bit simd instructions.
\newblock In \emph{2019 International Conference on Information and Communication Technology Convergence (ICTC)}, pp.\  1099--1103. IEEE, 2019.

\bibitem[Spring \& Shrivastava(2017)Spring and Shrivastava]{hashing_dl}
Spring, R. and Shrivastava, A.
\newblock Scalable and sustainable deep learning via randomized hashing.
\newblock In \emph{Proceedings of the 23rd ACM SIGKDD International Conference on Knowledge Discovery and Data Mining}, pp.\  445--454, 2017.

\bibitem[Sun et~al.(2019)Sun, Agostini, Dong, and Kaeli]{sun2019summarizing}
Sun, Y., Agostini, N.~B., Dong, S., and Kaeli, D.
\newblock Summarizing cpu and gpu design trends with product data.
\newblock \emph{arXiv preprint arXiv:1911.11313}, 2019.

\bibitem[Sung et~al.(2023)Sung, Hur, Kim, Ha, Oh, and Ro]{sung2023mad}
Sung, S., Hur, S., Kim, S., Ha, D., Oh, Y., and Ro, W.~W.
\newblock Mad macce: Supporting multiply-add operations for democratizing matrix-multiplication accelerators.
\newblock In \emph{Proceedings of the 56th Annual IEEE/ACM International Symposium on Microarchitecture}, pp.\  367--379, 2023.

\bibitem[Thirunavukarasu et~al.(2023)Thirunavukarasu, Ting, Elangovan, Gutierrez, Tan, and Ting]{llm_in_med}
Thirunavukarasu, A.~J., Ting, D. S.~J., Elangovan, K., Gutierrez, L., Tan, T.~F., and Ting, D. S.~W.
\newblock Large language models in medicine.
\newblock \emph{Nature medicine}, 29\penalty0 (8):\penalty0 1930--1940, 2023.

\bibitem[Touvron et~al.(2023)Touvron, Martin, Stone, Albert, Almahairi, Babaei, Bashlykov, Batra, Bhargava, Bhosale, et~al.]{llama2}
Touvron, H., Martin, L., Stone, K., Albert, P., Almahairi, A., Babaei, Y., Bashlykov, N., Batra, S., Bhargava, P., Bhosale, S., et~al.
\newblock Llama 2: Open foundation and fine-tuned chat models.
\newblock \emph{arXiv preprint arXiv:2307.09288}, 2023.

\bibitem[Tow et~al.()Tow, Bellagente, Mahan, and Riquelme]{StableLM-3B}
Tow, J., Bellagente, M., Mahan, D., and Riquelme, C.
\newblock Stablelm 3b 4e1t.
\newblock URL \url{[https://huggingface.co/stabilityai/stablelm-3b-4e1t](https://huggingface.co/stabilityai/stablelm-3b-4e1t)}.

\bibitem[Vaswani et~al.(2017)Vaswani, Shazeer, Parmar, Uszkoreit, Jones, Gomez, Kaiser, and Polosukhin]{attention}
Vaswani, A., Shazeer, N., Parmar, N., Uszkoreit, J., Jones, L., Gomez, A.~N., Kaiser, {\L}., and Polosukhin, I.
\newblock Attention is all you need.
\newblock \emph{Advances in neural information processing systems}, 30, 2017.

\bibitem[Wang et~al.(2020)Wang, Li, Khabsa, Fang, and Ma]{linformer}
Wang, S., Li, B.~Z., Khabsa, M., Fang, H., and Ma, H.
\newblock Linformer: Self-attention with linear complexity.
\newblock \emph{arXiv preprint arXiv:2006.04768}, 2020.

\bibitem[Wei et~al.(2022)Wei, Tay, Bommasani, Raffel, Zoph, Borgeaud, Yogatama, Bosma, Zhou, Metzler, et~al.]{emergent_abilities}
Wei, J., Tay, Y., Bommasani, R., Raffel, C., Zoph, B., Borgeaud, S., Yogatama, D., Bosma, M., Zhou, D., Metzler, D., et~al.
\newblock Emergent abilities of large language models.
\newblock \emph{arXiv preprint arXiv:2206.07682}, 2022.

\bibitem[Wold et~al.(1987)Wold, Esbensen, and Geladi]{wold1987principal}
Wold, S., Esbensen, K., and Geladi, P.
\newblock Principal component analysis.
\newblock \emph{Chemometrics and intelligent laboratory systems}, 2\penalty0 (1-3):\penalty0 37--52, 1987.

\bibitem[Xiao et~al.(2021)Xiao, Hu, Liu, Tu, and Sun]{xiao2021lawformer}
Xiao, C., Hu, X., Liu, Z., Tu, C., and Sun, M.
\newblock Lawformer: A pre-trained language model for chinese legal long documents.
\newblock \emph{AI Open}, 2:\penalty0 79--84, 2021.

\bibitem[Xiong et~al.(2021)Xiong, Zeng, Chakraborty, Tan, Fung, Li, and Singh]{nystromformer}
Xiong, Y., Zeng, Z., Chakraborty, R., Tan, M., Fung, G., Li, Y., and Singh, V.
\newblock Nystr{\"o}mformer: A nystr{\"o}m-based algorithm for approximating self-attention.
\newblock In \emph{Proceedings of the AAAI Conference on Artificial Intelligence}, volume~35, pp.\  14138--14148, 2021.

\bibitem[Zaheer et~al.(2020)Zaheer, Guruganesh, Dubey, Ainslie, Alberti, Ontanon, Pham, Ravula, Wang, Yang, et~al.]{bigbird}
Zaheer, M., Guruganesh, G., Dubey, K.~A., Ainslie, J., Alberti, C., Ontanon, S., Pham, P., Ravula, A., Wang, Q., Yang, L., et~al.
\newblock Big bird: Transformers for longer sequences.
\newblock \emph{Advances in neural information processing systems}, 33:\penalty0 17283--17297, 2020.

\bibitem[Zhang et~al.(2023{\natexlab{a}})Zhang, Ning, Prabhakar, and Wentzlaff]{zhang2023hardware}
Zhang, H., Ning, A., Prabhakar, R., and Wentzlaff, D.
\newblock A hardware evaluation framework for large language model inference.
\newblock \emph{arXiv preprint arXiv:2312.03134}, 2023{\natexlab{a}}.

\bibitem[Zhang et~al.(2019)Zhang, Yan, Lin, Zhao, and Peng]{zhang2019high}
Zhang, W., Yan, Z., Lin, Y., Zhao, C., and Peng, L.
\newblock A high throughput b+ tree for simd architectures.
\newblock \emph{IEEE Transactions on Parallel and Distributed Systems}, 31\penalty0 (3):\penalty0 707--720, 2019.

\bibitem[Zhang et~al.(2023{\natexlab{b}})Zhang, Sheng, Zhou, Chen, Zheng, Cai, Song, Tian, R{\'e}, Barrett, et~al.]{h2o}
Zhang, Z., Sheng, Y., Zhou, T., Chen, T., Zheng, L., Cai, R., Song, Z., Tian, Y., R{\'e}, C., Barrett, C., et~al.
\newblock H $ \_2 $ o: Heavy-hitter oracle for efficient generative inference of large language models.
\newblock \emph{arXiv preprint arXiv:2306.14048}, 2023{\natexlab{b}}.

\end{thebibliography}
